\documentclass[lettersize,journal]{IEEEtran}
\usepackage{amsmath,amsfonts}
\usepackage{algorithmic}
\usepackage{algorithm}
\usepackage{array}
\usepackage[caption=false,font=normalsize,labelfont=sf,textfont=sf]{subfig}
\usepackage{textcomp}
\usepackage{stfloats}
\usepackage{url}
\usepackage{verbatim}
\usepackage{booktabs}
\usepackage{multirow}
\usepackage{multicol}
\usepackage{cite}
\usepackage{color}
\usepackage{graphicx}
\usepackage{cite}
\begin{document}

\title{Perceptual MAE for Image Manipulation Localization: A High-level Vision Learner Focusing on Low-level Features}

\author{Xiaochen Ma\textsuperscript{1}, Jizhe Zhou\textsuperscript{1}, Xiong Xu\textsuperscript{2} , Zhuohang Jiang\textsuperscript{1}, Chi-Man Pun\textsuperscript{3}, \textit{Senior Member, IEEE} \\  \vspace{0.3cm}
\textsuperscript{1} College of Computer Science, Sichuan University \\
\textsuperscript{2} Second Laboratory, The 10th Institute of China Electronics Technology Group \\
\textsuperscript{3} Computer and Information Science, Faculty of Science and Technology, University of Macau \\
} 



\maketitle

\begin{abstract}
Nowadays, multimedia forensics faces unprecedented challenges due to the rapid advancement of multimedia generation technology thereby making Image Manipulation Localization (IML) crucial in the pursuit of truth. The key to IML lies in revealing the artifacts or inconsistencies between the tampered and authentic areas, which are evident under pixel-level features. Consequently, existing studies treat IML as a low-level vision task, focusing on allocating tampered masks by crafting pixel-level features such as image RGB noises, edge signals, or high-frequency features. However, in practice, tampering commonly occurs at the object level, and different classes of objects have varying likelihoods of becoming targets of tampering. Therefore, object semantics are also vital in identifying the tampered areas in addition to pixel-level features. This necessitates IML models to carry out a semantic understanding of the entire image. In this paper, we reformulate the IML task as a high-level vision task that greatly benefits from low-level features. Based on such an interpretation, we propose a method to enhance the Masked Autoencoder (MAE) by incorporating high-resolution inputs and a perceptual loss supervision module, which is termed Perceptual MAE (PMAE). While MAE has demonstrated an impressive understanding of object semantics, PMAE can also compensate for low-level semantics with our proposed enhancements. Evidenced by extensive experiments, this paradigm effectively unites the low-level and high-level features of the IML task and outperforms state-of-the-art tampering localization methods on all five publicly available datasets.
\end{abstract}


\section{Introduction}
Recent advancements in large generative models, such as Stable Diffusion\cite{stable_diffusion_2022}, have significantly improved the quality and diversity of multimedia content enhancement results. However, this technique is not without its potential drawbacks. In contrast to previous methods that require professional knowledge for operation and merely yield plausible outputs, large generative models endow ordinary people with easy access to generate imperceptive image manipulation results. This ability to create high-quality manipulated images on a massive scale has unleashed the problem of image tampering. Therefore, it creates unprecedented challenges for multimedia forensics, particularly in Image Manipulation Localization (IML). Effective IML methods are urgently needed to mitigate the negative impact of tampered images, such as fake news, rumors, and misleading information. In short, IML methods have become essential today to safeguard against the jeopardizes caused by image tampering and ensure that multimedia content remains trustworthy and reliable. In the context of the growing interest in image manipulation detection/localization, recent submissions~\cite{UNET_IML_2023, Chain_IML_2022, msae_2022, serial_copymove_2021} reflect the active pursuit of advancements in this area.

\IEEEpubidadjcol

\begin{figure*}
  \includegraphics[width=\textwidth]{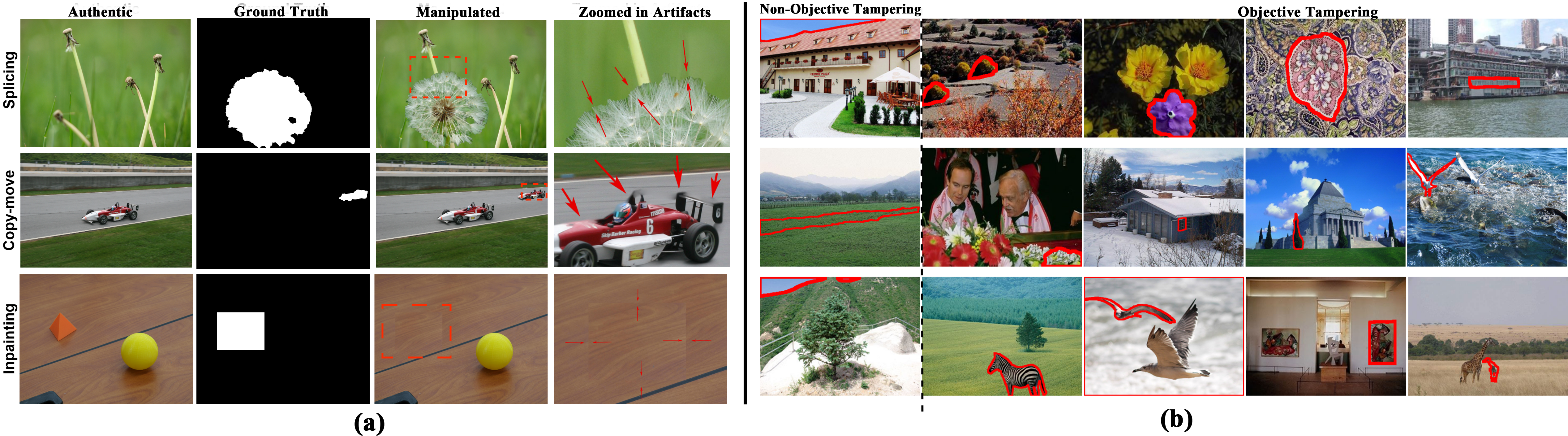}
  \caption{\textbf{ (a) Example of artifacts in three types of tampering.} The red dashed box in the third column represents the range of the zoomed-in area in the fourth column. Red arrows in the fourth column point to artifacts that are considered tampering traces. (b) Random samples from CASIAv2\cite{CASIA_2013} dataset, 80\% are object-related manipulation. The red line marks the boundary of the tampered area. The first column shows tampering that is unrelated to objects, while the other four columns show object-related tampering. For IML datasets, manipulation on objects is a common case, as it can more effectively confuse the semantics of the entire image.}
  \label{fig:artifacts}
\end{figure*}

From the perspective of image tampering, Figure \ref{fig:artifacts} (a) illustrates that existing techniques can be broadly classified into three categories\cite{Verdoliva_2020,Mantra_2019}: \textit{Splicing}(combining parts of different images to create a new one), \textit{Copy-move}(copying and pasting a region within the same image), and \textit{inpainting}(Remove and filling in an area with plausible content). Despite that large generative models can yield tampered results imperceptible to human eyes, each type of manipulation still leaves detectable traces at the pixel level. These traces manifest as inconsistencies between the tampered and authentic regions and are commonly referred to as \textit{artifacts}. Therefore, most existing image manipulation localization techniques treat IML as a \textbf{low-level vision} task that aims to capture the artifacts by extracting pixel-level features, such as the image RGB noises\cite{SRM_2018, Bayarconv_2018}, edge signals\cite{GSR_Net_2020, MVSS_2021}, or high-frequency features\cite{objectformer_2022,highpass_2019}. These low-level vision features are generally effective in revealing the artifacts and localizing the tampered regions. However, fully relying on low-level features leads existing IML models to suffer from low generalization ability and robustness. Therefore, constructing an approach that incorporates other manipulation traces is the key to improving localization accuracy and addressing the generalization and 
robustness limitations.

To achieve this, it is crucial to understand the characteristics and patterns of tampering. Typically, as shown in Figure \ref{fig:artifacts} (b), most tampering aims to deceive the audience by altering or confusing the semantics in images. As a result, tampering commonly occurs on objects rather than backgrounds within an image. Moreover, the likelihood of an object being targeted for tampering varies depending on its class and its contribution to the overall semantics of the image. For instance, humans and animals in the foreground are more likely targets for tampering than trees and mountains in the background. Therefore, we argue that understanding \textbf{high-level} visual information, like object-level semantics, could be useful in identifying manipulated regions and outlining suspicious areas completely. A recent study, ObjectFomer~\cite{objectformer_2022} using enhanced object proposals has experimentally supported this argument. However, high-level semantics alone are insufficient for generating tampering masks, as they lack comprehension of the detailed artifacts. Thus, a multi-level method fusing both low-level artifacts and high-level semantic features is the optimal solution for the IML task.

Hence, in this paper, we are the first ever to reformulate the IML task as
\textit{\textbf{a high-level vision task significantly benefits from low-level features.}} Such a character makes IML unique from any other tasks. 
To support the proposed argument, we searched among various self-supervised Vision Transformers~\cite{BEiT_2022, ibot_2021, DINO_2021, MAE_2022}, which are all highly proficient in learning high-level semantics, to identify potential candidates as our backbone. MAE~\cite{MAE_2022} stands out as the first method focused on pixel-level reconstruction, which could easily modified to fit in the low-level features. In contrast, others focus on reconstructing tokenized feature maps or complicated paradigms that are impossible to enhance with low-level information. As evidenced by experiments, MAE indeed outperforms other methods on IML tasks. Besides, IML often faces the dataset insufficiency problem. Common public IML datasets usually only have thousands and hundreds of images, which can not satisfy the appetite of a vanilla ViT. MAE pre-training is also powerful enough to help us overcome these issues.

To help the model focus on low-level information, We propose the \textbf{P}erceptual \textbf{M}asked \textbf{A}uto\textbf{e}ncoder (PMAE), a self-supervised module that enhances the model to cope with low-level artifacts in IML. Based on MAE, PMAE inherits its remarkable semantic comprehension and further enriches its learning ability of low-level visual features through a high-resolution encoder supervised by hierarchical perceptual loss. In the whole paradigm, we pre-trained a ViT encoder with MAE on large real-world datasets like ImageNet to learn object semantics. Then, during fine-tuning, we slightly modified the encoder with high-resolution patch embedding for tracing detailed features and tuned it on limited IML datasets with an IML segmentation branch and a PMAE reconstruction branch. Since these two branches share the same high-resolution encoder and optimize together, if the PMAE could reconstruct the low-level visual features well, then the latent representation learned in this process can also be effective for the segmentation branch. This paradigm allows the model to learn the high-level object semantics from a larger, more real-world sampled dataset and fully mine the tamper-related low-level visual features from the expensive and limited IML dataset. 


We follow a widely used evaluation protocol \cite{RGBN_2018, MVSS_2021, GSR_Net_2020, IML_ViT_2023} for IML to measure the performance and generalizability of our model. In detail, the model is trained on CASIAv2~\cite{CASIA_2013} datasets, then evaluates the metrics on smaller public datasets, including CASIAv1\cite{CASIA_2013}, Columbia~\cite{Columbia_2006}, COVERAGE~\cite{Coverage_2016} and NIST16~\cite{NIST16_2019}. The experimental results verify that PMAE has the ability to guide the model to outperform state-of-the-art ones on F1 score, AUC, and robustness. This plug-and-play module also provides the possibility of further exploration in conducting additional IML tasks with ViT. 

In summary, our contributions are as follows:
\begin{itemize}
    \item We revisit the essence of IML and reformulate the IML task as a high-level vision task that greatly benefits from low-level features. 
    \item According to our interpretation of the IML task, we establish the PMAE, a model with multi-level visual capturing ability that can effectively support image manipulation localization during fine-tuning.
    \item Extensive experiments show that PMAE outperforms state-of-the-art models on five public benchmark datasets, evaluated using $F_1$ scores and robustness metrics. This provides strong evidence to verify our interpretation of the IML task.

\end{itemize}

\section{Related works}

\textbf{Mask Image Modeling}
Taking inspiration from the success of masked language modeling in language tasks \cite{bert_2018}, masked image modeling (MIM) in the visual domain learns representations from images that are disrupted by masking. Several methods have achieved State-of-the-art results on downstream tasks. BEiT\cite{BEiT_2022} proposes to recover discrete visual labels, while SimMIM\cite{xie2022simmim} addresses the MIM task as a pixel-level reconstruction. In this work, we focus on MAE \cite{MAE_2022}, which proposes to use a high masking rate and non-arbitrary ViT decoder. A higher masking rate can increase the difficulty of reconstruction and force the model to focus on macro-level semantics. No structured modifications to the ViT encoder also facilitate our plug-and-play use of the recent new ViT algorithm. We will further discuss why we don't select other self-supervised ViT as the backbone in Section \ref{sec:mae_for_vit_encoder}.

\textbf{Image Manipulation Detection/Localization} In the early years, image manipulation detection usually focuses on single-type tampering, especially copy-move detectors like Dense-InceptionNet~\cite{e2e_copy_move_2019} and STRDNet~\cite{serial_copymove_2021} that identify potential copy-move forgery instances. Low-level visual features, such as noise, Sobel (edge detection), and high-pass filters, have shown excellent performance for specific types of tampering and become prevalent. After that, generic tampering detection by end-to-end deep learning methods became dominant, which is manipulation type-independent. Most of them combine the RGB view with other low-level vision views and become successful. 
RGB-N\cite{RGBN_2018} proposed the SRM filter to extract noise features and support the detection by Faster R-CNN-based network. The bayarConv filter proposed in Constrained CNN\cite{Bayarconv_2018} can also extract noise information for supporting classification. J. Bappy \textit{et al.}~\cite{bappy2019hybrid} employ a hybrid CNN-LSTM model that effectively
 classifies manipulated and non-manipulated regions.
Wu Yue \textit{et al.} firstly concatenates the feature maps from the SRM filter and the BayarConv together and uses VGG as the backbone to complete the segmentation of the manipulated area. Recent work of MVSS-Net\cite{MVSS_2021} and MVSS-Net++ \cite{mvsspp_2022} also utilizes BayarConv and combines it with sobel filters with edge detection ability by dual-attention. Objectformer\cite{objectformer_2022} uses Discrete Cosine Transform to acquire high-frequency features to obtain information that is difficult to get through RGB channels.
However, these features were initially designed for specific tampering methods and not generalized. We suggest that utilizing self-supervised methods to discover traces autonomously could be a better choice than these handcrafted feature extractors.

\section{Proposed Methods}
Our goal is to enhance MAE's understanding of low-level visual features, especially subtle traces related to tampering. Building upon MAE's inherent object-level reconstruction capability, we have devised during the finetuning stage an enhanced self-supervised method, Perceptual Masked Autoencoder(PMAE), to bolster the model's sensitivity to low-level artifacts. PMAE holds a significant advantage over the earlier hand-crafted filters, as it can learn the most significant features from the dataset by itself, rather than relying solely on narrow prior knowledge.

In this section, we introduce our whole training paradigm and the implementation detail of PMAE, which enhances MAE with high-resolution input and supervises it with a hierarchical masked perceptual loss.

\subsection{Overview of Training Paradigm}
\begin{figure}[h]
  \includegraphics[width=1.0\columnwidth]{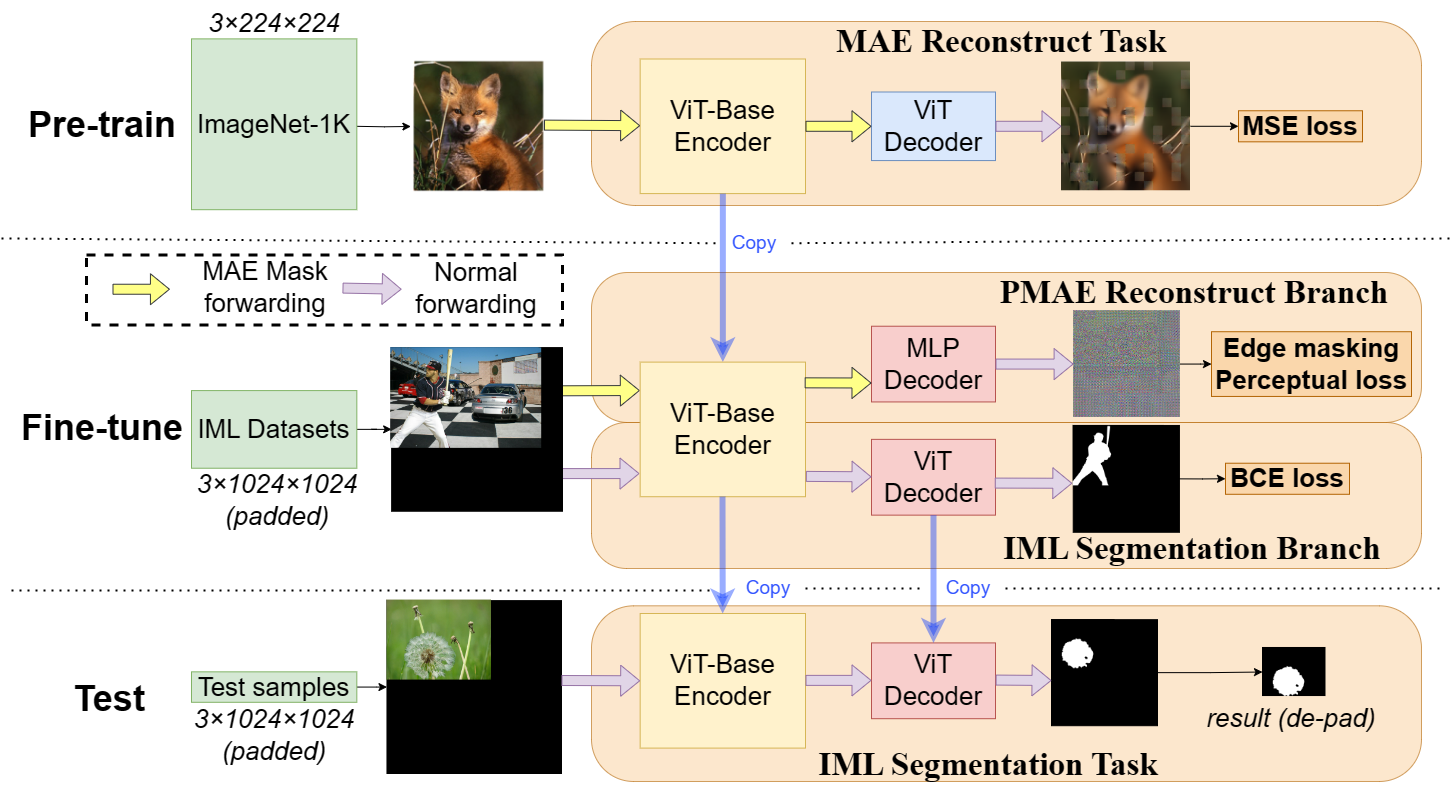}
  \caption{Diagram of the pre-train and fine-tune process of proposed paradigm.}
  \label{fig:paradigm}
\end{figure}

The widely adopted paradigm of pre-training, followed by fine-tuning, is utilized in our work. We commence with an MAE pre-training on the low-resolution ImageNet-1k dataset, imbuing the model with the semantics of common objects. Subsequently, as illustrated in Figure \ref{fig:paradigm}, we directly transfer the parameters of the pre-trained ViT encoder for our fine-tuning process on padded high-resolution IML datasets that keep the freshest artifacts to learn. Our fine-tuning involves two distinct tasks: the \textit{IML segmentation} and \textit{PMAE reconstruct} branch. The segmentation branch employs a deliberately uncomplicated structure to segment suspicious regions like IML-VIT~\cite{IML_ViT_2023}. In contrast, the PMAE branch informs the model with detailed distributions learned from IML datasets. These two branch shares the same ViT encoder and utilize their own decoder. We jointly optimize both branches by computing the gradient together. However, since the MAE masking strategy and segmentation forwarding provide different inputs for the ViT encoder, each image will pass through the ViT encoder twice.

In general, the optimization target of the IML segmentation branch can be formulated as:
\begin{equation}
  \underset{\theta_E, \theta_S}{ \arg\,\min }\, \mathcal{L}_{seg} \{ \mathcal{D}_{MLP}[ \mathcal{E} (X_p ; \theta_E) ; \theta_S ],  M_p\}
\end{equation}
while the PMAE reconstruct branch can be formulated as:
\begin{equation}
  \underset{\theta_E, \theta_R}{ \arg\,\min }\, \mathcal{L}_{rec} \{ \mathcal{D}_{ViT}[ \mathcal{E} (X_p' ; \theta_E) ; \theta_R ],  X_p\}
\end{equation}
Here, $\mathcal{L}$ refers to the loss functions for the respective branches, $\mathcal{D}$ represents the decoder, and $\mathcal{E}$ represents the encoder. All the $\theta$ refer to the corresponding model parameters. In these formulas, $X$ represents the distribution of the input images, $X'$ represents the distribution of the data after MAE random masking and $M$ represents the ground truth mask. The subscript $p$ denotes the zero-padding operation, which will be further explained in Section \ref{sec:highresolution}.

We do not perform independent pre-training for the PMAE branch because the high-quality public IML datasets are relatively small. Even a larger dataset in this field, CASIAv2, contains only 5063 tampered images and 7491 authentic images, which is significantly smaller than the 1.2 million images in ImageNet-1k. This dataset size discrepancy makes it challenging to determine whether the model has fully converged or overfitted. Because we lack appropriate indicators to monitor the reconstruction process. Therefore, we accompany the segmentation branch during training to evaluate the optimization process promptly by monitoring its performance on the test dataset.

\begin{figure*}[t]
  \includegraphics[width=1.0\textwidth]{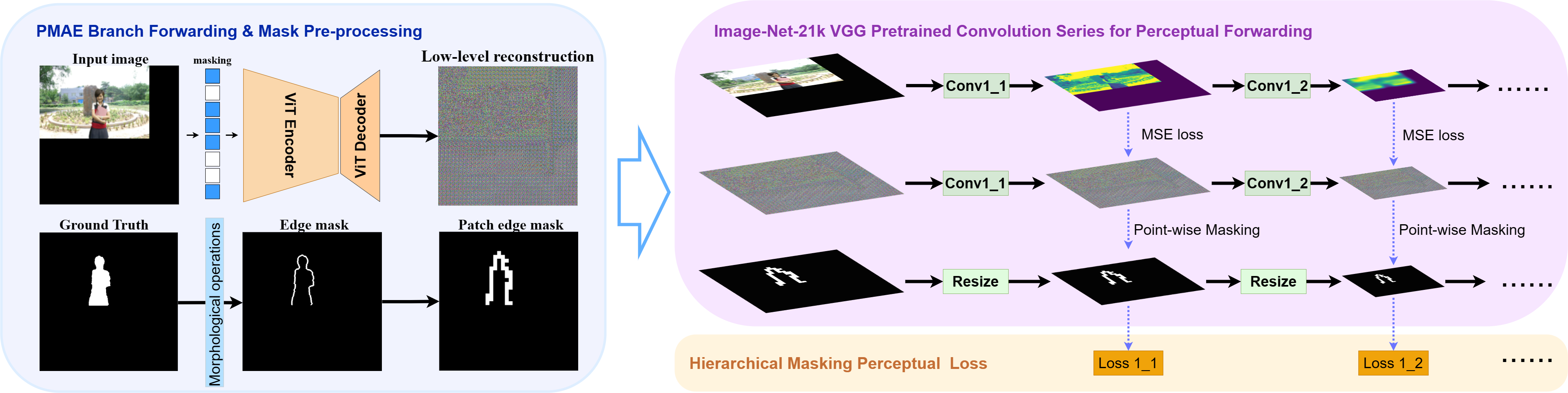}
  \caption{Overview of the Perceptual Masked Autoencoder (PMAE) reconstruction branch. Since our perceptual loss only computes with loss from shallow convolution layers and only focuses on the edge-related patches, reconstruction images often look like a noisy map. However, this indicates our method truly captures the low-level vision features from datasets  }
  \label{fig:PMAE}

\end{figure*}

\subsection{MAE Pre-training on ViT Encoder}
\label{sec:mae_for_vit_encoder}
Recently, self-supervised Vision Transformers like MAE~\cite{MAE_2022}, Beit~\cite{BEiT_2022}, iBOT~\cite{ibot_2021}, and DINO~\cite{DINO_2021} have been shown to have impressive performance in various downstream classification tasks, which also signifies their strong understanding of object-level semantics. However, we ultimately select MAE as our object semantic learner based on the following reasons:
(1) MAE stands out as the first method focused on pixel-level reconstruction, while others focus on reconstructing the tokenized (by methods like VAE~\cite{VAE_2013}) feature maps, thereby MAE is more competitive in tracing low-level information.
(2) In models like DINO, it is hard to design a structure that could effectively focus on low-level features and maintain its original transfer learning paradigm at the same time.
(3) MAE is very plain, with almost only one naive ViT, bringing two benefits: First, there is almost no need to introduce any additional modules. Second, the PMAE optimized for low-level traces can easily maintain almost the same pattern as MAE, making the model converge quickly. 
Furthermore, in Section \ref{sec:ablation}, we will demonstrate through experiments that MAE can indeed outperform other self-supervised methods.

\subsection{High-resolution ViT Encoder for Fine-tuning}
\label{sec:highresolution}
During fine-tuning, both the segmentation branch and the PMAE reconstruction branch require the ViT encoder to extract intricate details and artifacts from images as much as possible. To achieve this, it is essential to \textbf{preserve the original resolution} of each image to avoid downsampling that could potentially distort the artifacts. However, when computing images in parallel, all images within a batch must have the same resolution. To reconcile these competing demands, we adopt a novel approach. Rather than simply rescaling images to the same size, we pad the images and ground truth masks with zeros instead of resizing, then place the image on the top-left side to match a larger constant resolution. This strategy maintains crucial low-level visual information of each image, allowing the model to explore better features instead of depending on handcrafted prior knowledge. To implement this approach, we first adjust the patch embedding dimensions of the ViT encoder to a larger scale. However, this modification significantly increases the computing complexity. To balance it, we adopt a technique inspired by previous works\cite{MViTv2_2022,Benchmark_ViT_2021}, which periodically replaces part of the global attention blocks in ViT with windowed attention blocks. This method ensures global information propagation while reducing the computational cost.

More specifically, we represent input images as $X\in \mathbb{R}^{3\times h \times w }$, and ground truth masks as $M\in \mathbb{R}^{1\times h \times w }$, where $h$ and $w$ correspond to the height and width of the image, respectively. We then pad them to $X_p \in \mathbb{R}^{ 3 \times H \times W }$ and $M_p \in \mathbb{R}^{ 1 \times H \times W }$. Balance with computational cost and the resolution of common IML datasets(see in Table \ref{tab:datasets}), we take $H=W=1024$ as consts in our implementation. Then $X_p$ is passed into the windowed ViT-B encoder with 12 layers, with a complete global attention block retained every 3 layers. 
 
\subsection{MLP Segmentation Branch}
\begin{figure}[t]
  \includegraphics[width=1.0\columnwidth]{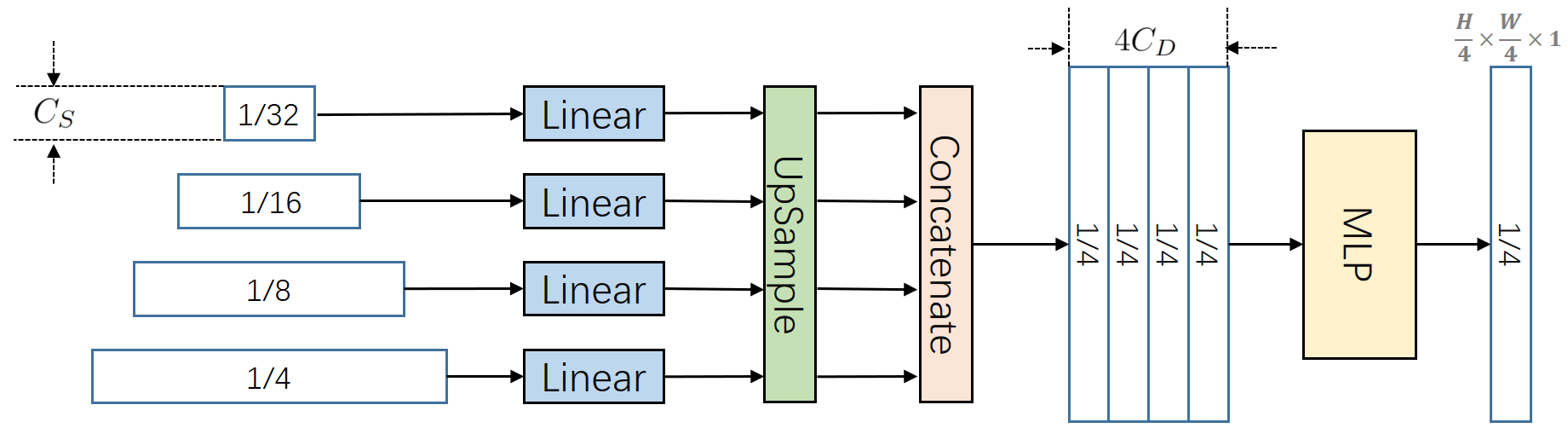}
  \caption{Diagrams of the MLP decoder. White rectangles on the left represent the output of the Simple Feature Pyramid. Fractions denote for the resolution compared to padded images.}
  \label{fig:MLP}
\end{figure}

\textbf{Simple Feature Pyramid}
To incorporate multi-scale supervision that is efficient for segmentation, we employ a feature pyramid network after the ViT encoder, following the approach proposed in ViTDet~\cite{ViTDet_2022}. This method uses the single output feature map $G_e = \mathcal{E} (X_p, ; \theta_E)$ from ViT and applies a series of convolutional and deconvolutional layers to upsample and downsample the feature map to obtain multi-scale feature maps $F_i$, where $i \in {1,2,3,4}$:
\begin{equation}
F_i = \mathcal{C}_i(G_e) \in \mathbb{R}^{\frac{H}{2^{i+2}}\times \frac{W}{2^{i+2}} \times C_{S}}
\end{equation}
Here, $\mathcal{C}_i$ denotes the convolution series, and $C_S$ represents the output channel dimension for each layer in the feature pyramid. Notably, this multi-scale method does not alter the base structure of ViT, allowing for the easy introduction of recent advanced algorithms like MAE to the backbone.

\textbf{Lightweight Prediction Head}
To reduce memory consumption while demonstrating the improvements from the advanced design of the ViT encoder and PMAE supervise, we aim to apply a lightweight network for the final prediction. To this end, as shown in Figure \ref{fig:MLP}, we adopt the decoder design from SegFormer~\cite{segformer_2021}, which outputs a smaller predicted mask $\hat{M}$ with a resolution of $1 \times \frac{H}{4}\times \frac{W}{4} $, which can effectively reduce computational complexity. This lightweight All-MLP decoder first applies a linear layer to unify the channel dimension and then upsamples all features to the same resolution of $ C_D \times \frac{H}{4}\times \frac{W}{4}$ using bilinear interpolation. Subsequently, we concatenate all the features and apply a series of linear layers to fuse them and make the final prediction. The prediction head can be expressed as follows:

\begin{align}
  P = &\, \mathcal{D}_{MLP}[ \mathcal{E} (X_p ; \theta_E) ; \theta_S ] \\
    = &\, MLP\{ \odot_i (W_iF_i+b_i) \}  \in \mathbb{R} ^{\frac{H}{4}\times \frac{W}{4} \times 1} 
\end{align}

Here, $P$ represents the predicted probability map for the manipulated area; $\odot$ denotes for concatenation operation, and $MLP$ refers to an MLP module.
The loss for the segmentation branch is computed with binary cross entropy loss function: $\mathcal{L}_{seg} = \mathcal{L}_{BCE}(P, M_p)$.

\subsection{PMAE Reconstruction Branch}
The Perceptual Masked Autoencoder(PMAE) first obtains tokens encoded by a ViT backbone from high-resolution padded images that have undergone random masking. Most of the settings in this process are the same as the original MAE, while the only exception is the number of patches significantly increased due to the high resolution of the input images. Then, we apply a series of Vision Transformer layers with full global attention as the decoder. Finally, a full-connected layer decodes the tokens back to an RGB image as the final reconstruction $R= \mathcal{D}_{ViT}[ \mathcal{E} (X_p' ; \theta_E) ; \theta_R ] \in \mathbb{R} ^{3\times H\times W}$.
Since we want to guide the model to learn low-level visual features related to tampering, and an important prior knowledge in IML is that tampering traces are largely distributed around the tampered area, we do not simply calculate perceptual loss\cite{perceptual_loss_2016} between the reconstructed image and the input image. Instead, we use a hierarchical masked perceptual loss to supervise the reconstructed image.

To implement this approach, we start by generating an \textit{edge mask} based on the ground truth using morphology operations. Next, we utilize this mask to further generate a \textit{patch edge mask}. As depicted in Figure \ref{fig:PMAE}, we first divide the \textit{edge mask} into patches, and if any pixel in a patch is equal to 1, we consider the entire region of the \textit{patch edge mask}  that corresponds to that patch to be 1. 
Finally, when calculating the perceptual loss, we apply a point-wise product between the \textit{patch edge mask} and the input feature map, as well as the reconstructed feature map after each convolution layer, before computing the MSE loss. This ensures that the model focuses only on the areas related to the tampered artifact.

The model we use to generate perceptual feature maps is a VGG\cite{vgg_2015} network on ImageNet-21k. Since our masking perceptual strategy has damaged the object-level semantics, we only adopt the masked perceptual features from coming layers: \textit{conv1\_2, conv2\_2, conv3\_2}, which are all shallow layers of VGG that only capture the low-level features, this selection following the original paper of perceptual loss\cite{perceptual_loss_2016} for low-level vision task. The final \textit{hierarchical masking perceptual loss} in our implementation can be formulated as:
\begin{align}
\mathcal{L}_{rec} &= \mathcal{L}_{rec} (R, X_p) \\
 &=\mathcal{L}_{rec}\{ \mathcal{D}_{ViT}[ \mathcal{E} (X_p' ; \theta_E) ; \theta_R ],  X_p\} \\
&= Loss_{1\_2} + Loss_{2\_2} + Loss_{3\_2}
\end{align}
where $Loss_{m\_n}$ denotes the single layer hierarchical masking perceptual loss in Figure \ref{fig:PMAE}. 

In summary, all the modifications on PAME compared to MAE aim to mine more low-level visual features during fine-tuning.

\subsection{Combined Loss} 
Even though the segmentation and the reconstruction branch have similar optimization goals, it can seriously affect the model's convergence performance if they are not balanced well. We formulate the final loss $\lambda$ and seek optimal $\lambda$:
\begin{equation}
  \mathcal{L} = \mathcal{L}_{seg} + \lambda \cdot \mathcal{L}_{rec}
\end{equation}
One significant factor is that the predicted values and ground truth values in the segmentation branch are always within the range of $[0,1]$, and the distance between them is not far. In contrast, the perceptual loss in the reconstruction branch is calculated based on the feature map of the middle layer of the VGG network, without activation functions like softmax that can normalize the values, which leads to a long distance between the reconstructed feature maps and the ground truths.
On average, the value of segmentation loss falls around 1e-2 to 1e-3, while the reconstruction loss falls around 10 to 100. We tested for $\lambda \in \{1, 0.1, 0.01, 0.001\}$ and finally selected the optimal value as 0.01.

\section{Experiments}

\subsection{Evaluation Barrier}
Despite the proliferation of the SoTA models in recent research, achieving equitable comparisons remains intricate. This difficulty stems partly from the absence of publicly accessible code and training methodologies for these models~\cite{SPAN_2020,objectformer_2022}. Furthermore, many studies rely on extensive synthesized datasets that are not available to the research community~\cite{Mantra_2019,MM-Net}. Hence, we advocate for the community's embrace of open-source practices and emphasize the necessity to evaluate dataset generation strategies independently from model performance. These measures are pivotal in ensuring equity and fostering continued progress in this domain.
\subsection{Experimental Setup} 

\textbf{Datasets}
To ensure a fair comparison with state-of-the-art methods in image tampering localization, we adopt a commonly used protocol~\cite{MVSS_2021, GSR_Net_2020, SPAN_2020} for our evaluation. We first train our model on the CASIAv2~\cite{CASIA_2013} dataset and then evaluate its performance on smaller public datasets including CASIAv1~\cite{CASIA_2013}, NIST16~\cite{NIST16_2019}, COVERAGE~\cite{Coverage_2016}, Columbia~\cite{Columbia_2006}, and Defacto~\cite{defacto_2019}, details can found in Table \ref{tab:datasets}. However, we note that the Defacto dataset does not contain authentic images as negative examples. To overcome this limitation, we follow the approach of MVSS-Net\cite{MVSS_2021} and randomly select 6000 untouched images from MS-COCO~\cite{mscoco_2014}. These images are combined with 6000 images from the Defacto dataset to create a validation set, called Defacto-12k.

\begin{table}[ht]
  \centering
 \caption{\textbf{Details of six datasets in our experiments.} }
  \label{tab:datasets}
  \resizebox{1.0\columnwidth}{!}{
  \begin{tabular}{@{}lllllllll@{}}
  \toprule[2pt]
  \multicolumn{1}{c}{\multirow{2}{*}{\textbf{Usage}}} &
    \multicolumn{1}{c}{\multirow{2}{*}{\textbf{Dataset}}} &
    \multicolumn{2}{c}{\textbf{Type}} &
    \multicolumn{3}{c}{\textbf{Manipulation Type}} &
    \multicolumn{2}{c}{\textbf{Resolution}} \\ \cmidrule(l){3-9} 
  \multicolumn{1}{c}{} &
    \multicolumn{1}{c}{} &
    \multicolumn{1}{c}{\textbf{Auth}} &
    \multicolumn{1}{c}{\textbf{Mani}} &
    \multicolumn{1}{c}{\textbf{copymv}} &
    \multicolumn{1}{c}{\textbf{spli}} &
    \multicolumn{1}{c}{\textbf{inpa}} &
    \multicolumn{1}{c}{\textbf{min}} &
    \multicolumn{1}{c}{\textbf{max}} \\ \midrule
  Train                 & CASIAv2~\cite{CASIA_2013}    & 7491 & 5063 & 3235 & 1828 & 0    & 240 & 800  \\\midrule[0.5pt]
  \multirow{5}{*}{Test} & CASIAv1~\cite{CASIA_2013}     & 800  & 920  & 459  & 461  & 0    & 256 & 384  \\
                        & NIST16~\cite{NIST16_2019}      & 0    & 564  & 68   & 288  & 208  & 480 & \textbf{5616} \\
                        & COVERAGE~\cite{Coverage_2016}    & 100  & 100  & 100  & 0    & 0    & 158 & 572  \\
                        & Defacto-12k~\cite{defacto_2019} & 6000 & 6000 & 2000 & 2000 & 2000 & 120 & 640  \\
                        & Columbia~\cite{Columbia_2006}    & 183  & 180  & 0    & 180  & 0    & 568 & \textbf{1152} \\ \bottomrule[2pt]
  \end{tabular}
  }
  \end{table}

\textbf{Evaluation Criteria}  
We assessed the effectiveness of our model in localizing image manipulations using two widely adopted metrics: the pixel-level $F_1$ measure and the pixel-level AUC measure. However, the AUC measure can be affected by imbalanced data, which is typically the case in IML datasets that contain more negative pixels. This can result in an overestimation of the model's performance. Therefore, to provide a more meaningful and practical evaluation of our model's performance, we focus on reporting the pixel-level $F_1$ score using a uniform threshold of $0.5$. This scoring approach is less susceptible to the influence of imbalanced data and is widely used as a robust metric for evaluating the effectiveness of image manipulation localization models.

\textbf{Implementation Details}
Our model is implemented with PyTorch and trained on NVIDIA RTX 3090 GPUs for 200 epochs with a batch size of 1. 
We initialized the ViT-B backbone with MAE pre-trained weights on ImageNet-1k and used the AdamW optimizer~\cite{AdamW_2019} with a base learning rate of 1e-4. We employed a cosine decay strategy~\cite{cosine_decay_2017} to schedule the learning rate. We applied the early stop technique during training to prevent overfitting. The PMAE branch and predict branch are sequentially processed within a batch. Each branch performs an independent back propagation step, thereby not concurrently occupying GPU memory.
To prepare the images for training, we added top-left zero-padding to all images (except those that exceeded the limit) to achieve a resolution of 1024x1024. Images with longer edges that exceeded the size limit were resized its longer side to 1024 while maintaining their aspect ratio. We applied standard data augmentation techniques such as rescaling, flipping, blurring, rotation, and basic manipulations (e.g., randomly copying, moving, or inpainting rectangular areas within a single image) during training.

In terms of inference, the computational cost is significantly reduced as only the segmentation branch is utilized. A batch size of 4 is employed, resulting in an approximate GPU memory consumption of 11GB, which can meet the requirements of most graphics cards. The average inference time for a single image on a 3090 GPU is approximately 0.6 seconds.
\subsection{Ablation Study}
\label{sec:ablation}

\begin{table*}[ht]
  \centering
  \caption{Ablation study of PAME on 4 public datasets, evaluated with pixel-level F1 score and pixel-level AUC. Best number per column is shown in bold. Models are all trained on CASIAv2 datasets.}
  \label{tab:ablation}
  
  \resizebox{1.0\textwidth}{!}{
    \begin{tabular}{@{}c|c|ccc|cccccccccc@{}}
      \toprule[2pt]
      \multirow{2}{*}{\textbf{Goal}} &
        \multirow{2}{*}{\textbf{Init method}} &
        \multicolumn{3}{c|}{\textbf{Components}} &
        \multicolumn{2}{c}{\textbf{CASIAv1}} &
        \multicolumn{2}{c}{\textbf{Coverage}} &
        \multicolumn{2}{c}{\textbf{Columbia}} &
        \multicolumn{2}{c}{\textbf{NIST16}} &
        \multicolumn{2}{c}{\textbf{MEAN}} \\ \cmidrule(l){3-15} 
       &
         &
        H-Reso &
        S-FPN &
        PMAE &
        F1 &
        AUC &
        F1 &
        AUC &
        F1 &
        AUC &
        F1 &
        AUC &
        F1 &
        AUC \\ \midrule
      \multirow{2}{*}{w/o MAE} &
        Xavier &
        + &
        + &
        - &
        0.1035 &
        - &
        0.0439 &
        - &
        0.0744 &
        - &
        0.0632 &
        - &
        0.0713 &
        - \\
       &
        ViT-B ImNet-21k &
        + &
        + &
        - &
        0.5114 &
        - &
        0.1854 &
        - &
        0.2287 &
        - &
        0.1811 &
        - &
        0.2767 &
        - \\
      w/o H-Reso &
        MAE ImNet-1k &
        - &
        + &
        - &
        0.5061 &
        0.8166 &
        0.2324 &
        0.8250 &
        0.5409 &
        0.8420 &
        0.2987 &
        0.8212 &
        0.3945 &
        0.8262 \\
      w/o S-FPN &
        MAE ImNet-1k &
        + &
        - &
        - &
        0.5996 &
        0.8627 &
        \textbf{0.4457} &
        0.8352 &
        0.6125 &
        0.8350 &
        0.1841 &
        0.6767 &
        0.4605 &
        0.8024 \\

      w/o PMAE &
        MAE ImNet-1k &
        + &
        + &
        - &
        0.5886 &
        0.8668 &
        0.3277 &
        0.8131 &
        0.7445 &
        0.9076 &
        0.2993 &
        0.7706 &
        0.4900 &
        0.8395 \\
      Full Setup &
        MAE ImNet-1k &
        + &
        + &
        + &
        \textbf{0.6267} &
        \textbf{0.9366} &
        0.3583 &
        \textbf{0.9285} &
        \textbf{0.7574} &
        \textbf{0.9298} &
        \textbf{0.3137} &
        \textbf{0.8313} &
        \textbf{0.5140} &
        \textbf{0.9065}
         \\ \bottomrule[2pt]
      \end{tabular}
  }

\end{table*}

\begin{table*}[t]
  \centering
  \caption{PAME compared with the state-of-the-art. The best scores are shown in bold.}
  \label{tab:sota}
  \resizebox{1.82\columnwidth}{!}{
  \begin{tabular}{@{}lcccccc@{}}
  \toprule[2pt]
  \multirow{2}{*}{Method} & \multicolumn{6}{c}{Pixel-level $F1$ score}                          \\ \cmidrule(l){2-7} 
                          & CASIAv1\cite{CASIA_2013} & Columbia\cite{Columbia_2006} & NIST16\cite{NIST16_2019}  & Coverage\cite{Coverage_2016}       & Defacto-12k\cite{defacto_2019} & MEAN  \\ \midrule
  HP-FCN, ICCV19\cite{HPFCN_2019}                  & 0.154   & 0.067    & 0.121 & 0.003          & 0.055       & 0.080 \\
  ManTra-Net, CVPR19\cite{Mantra_2019}             & 0.155   & 0.364    & 0     & 0.286          & 0.155       & 0.192 \\
  CR-CNN, ICME20\cite{CR_CNN_2020}                 & 0.405   & 0.436    & 0.238 & 0.291          & 0.132       & 0.300 \\
  GSR-Net, AAAI20\cite{GSR_Net_2020}          & 0.387   & 0.613    & 0.283 & 0.285          & 0.051       & 0.324 \\
  MVSS-Net, ICCV21\cite{MVSS_2021}        & 0.452   & 0.638    & 0.292 & 0.453          & 0.137       & 0.394 \\
  MVSS-Net(re-trained)      & 0.435   & 0.441    & 0.203 & 0.329          & 0.105       & 0.303 \\
  MVSS-Net++, TPAMI22\cite{mvsspp_2022}              & 0.513   & 0.660     & 0.304 & \textbf{0.482} & 0.095       & 0.411 \\ \midrule
  \textit{PMAE (ours)} & \textbf{0.688} & \textbf{0.860} & \textbf{0.311} & 0.473 & \textbf{0.177} & \textbf{0.502} \\ \bottomrule[2pt]
  \end{tabular}
  }

  \end{table*}

In this section, we perform ablation experiments to systematically analyze the contribution of each component in our proposed PMAE paradigm to the overall performance. Specifically, we investigate the impact of removing the following components: (1) MAE: initialize the ViT with Xavier init\cite{Xavier_2010}, ImageNet-21k classification, and other self-supervised strategies; (2) High-resolution: reduce the input resolution of fine-tuning by resizing all the images and masks to 512×512; (3) Simple Feature Pyramid: replace this multi-scale structure with a series of plain convolution layers; (4) PMAE branch: remove PMAE branch and related loss functions.

  
  The quantitative results on 4 widely used public IML datasets are presented in Table \ref{tab:ablation}. We report the pixel-level F1 score and pixel-level AUC as metrics.   Our ablation experiments demonstrate that each component contributes more or less to the overall performance of the model.
  
\begin{figure}[h] 
    \centering
    \includegraphics[width=0.45\columnwidth]{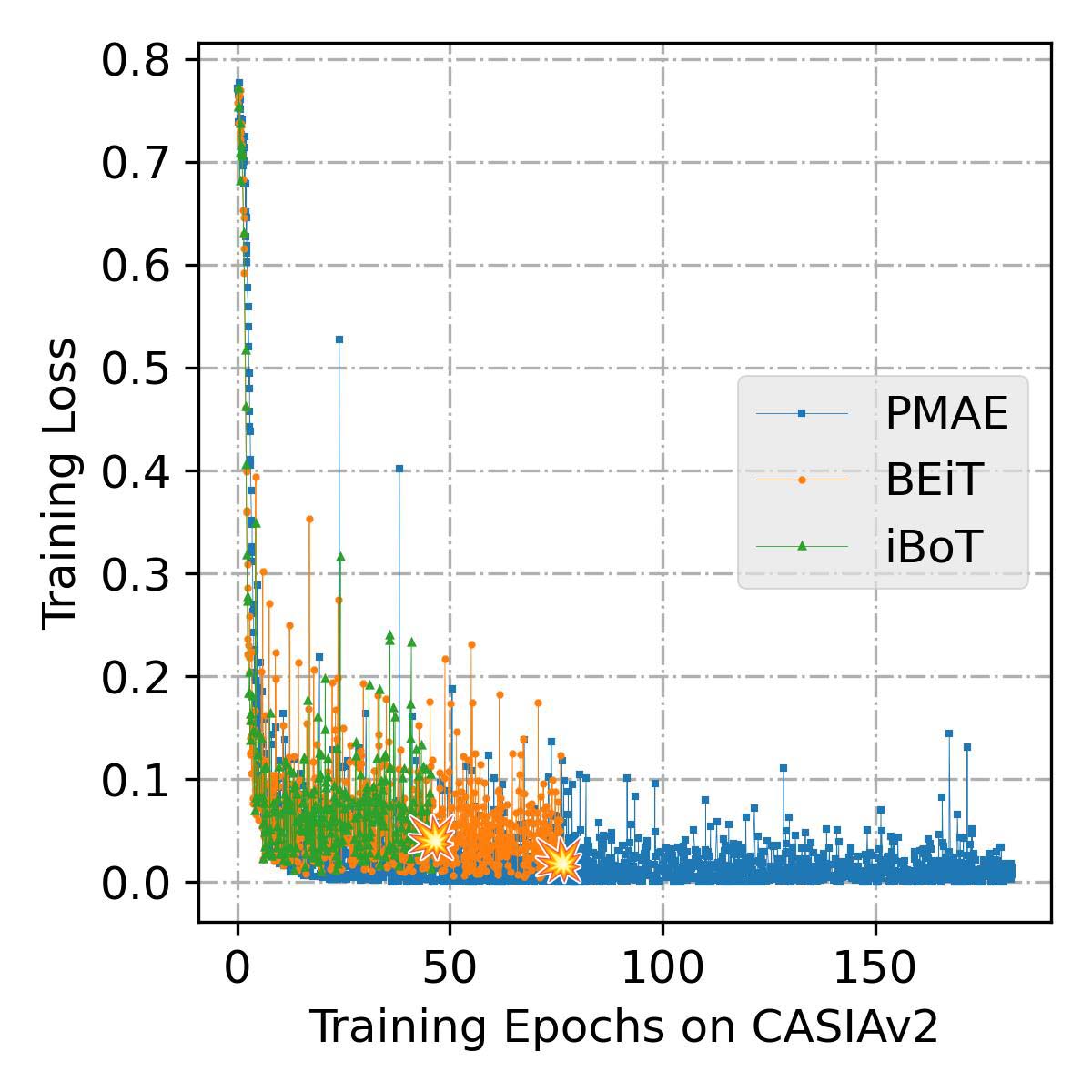} 
        \includegraphics[width=0.45\columnwidth]{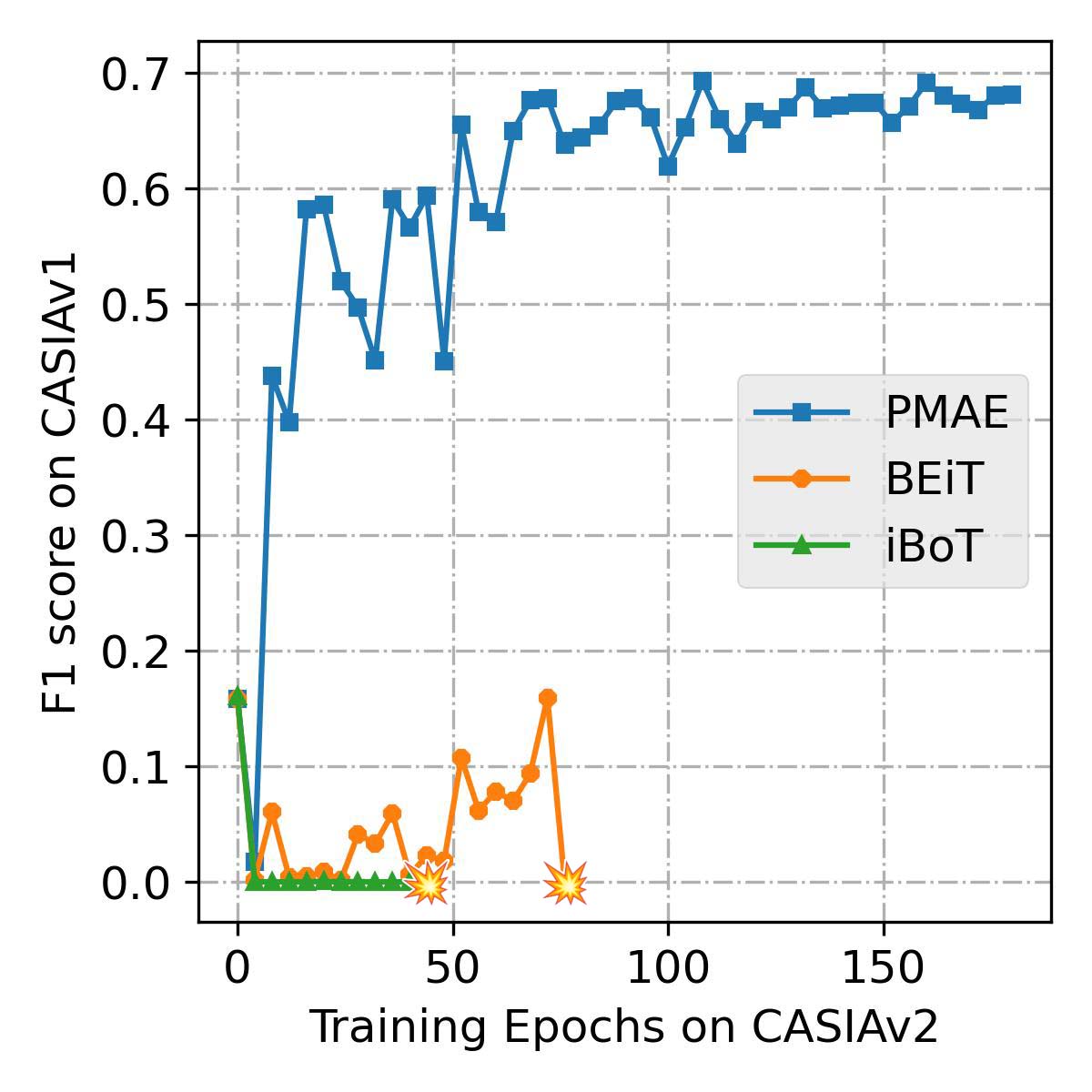} 
    \caption{Training loss and F1 score on test dataset for self-supervised pre-training algorithms. ``Explosion'' stickers represent gradient explosion/vanishing problem with the loss becoming NaN(not a number). }
    \label{fig:ssl_vit_ablation}
\end{figure}

\textbf{MAE init} 
As depicted in Figure \ref{fig:ssl_vit_ablation}, we have tested with various self-supervised ViT strategies to initialize our model. Except for MAE, they all experienced poor ability on IML tasks and could not converge eventually. We believe that the pixel-level reconstruction task designed for MAE demonstrates the capability to extract low-level semantics effectively, aiding the model in convergence. In addition, as shown in Table ~\ref{tab:ablation}, the model trained using the Xavier initialization method also encounters convergence issues, whereas traditional classification pre-training struggles to generalize effectively on non-homologous datasets. However, other settings with MAE initialization show at least a 21.8\% improvement in the average F1 score and converge rapidly, indicating that using the MAE init with objective semantics and low-level vision capacity can greatly aid the convergence and alleviate overfitting.

  \textbf{High-resolution} 
  The model performance after removing the high resolution is significantly reduced except for NIST16. For NIST16, because more of the images are much larger than 1024×1024 resolution, there is already a large amount of low-level features destroyed by downsampling when preprocessing. So it can be considered that the decision is still mainly supported by object semantics on this dataset, so there is not much change compared to others. This also indirectly proves that multi-level visual feature is indeed meaningful to solve the IML problem effectively.

  \textbf{Simple Feature Pyramid} 
  Performance on the COVERAGE dataset appears to be better without the simple feature pyramid compared to the Full setup. However, as indicated in Table \ref{tab:datasets}, the limited COVERAGE dataset only has 100 manipulated images, and the tampering type is restricted to copy-move only. We contend that achieving good performance on this dataset alone may indicate overfitting. In contrast, effectiveness on larger, more diverse datasets that are more practical and valuable. Thus, we argue that the simple feature pyramid generalizes the model on new and varied tampering scenarios.

  \textbf{PMAE} The PMAE module is an essential component of our model, as demonstrated by the ablation study results. The average F1 score increased by 4.89\% with the PMAE branch, confirming that the PMAE module provides valuable low-level visual information for the model to identify tampered regions accurately.

\begin{figure}[t]
  \includegraphics[width=0.95\columnwidth]{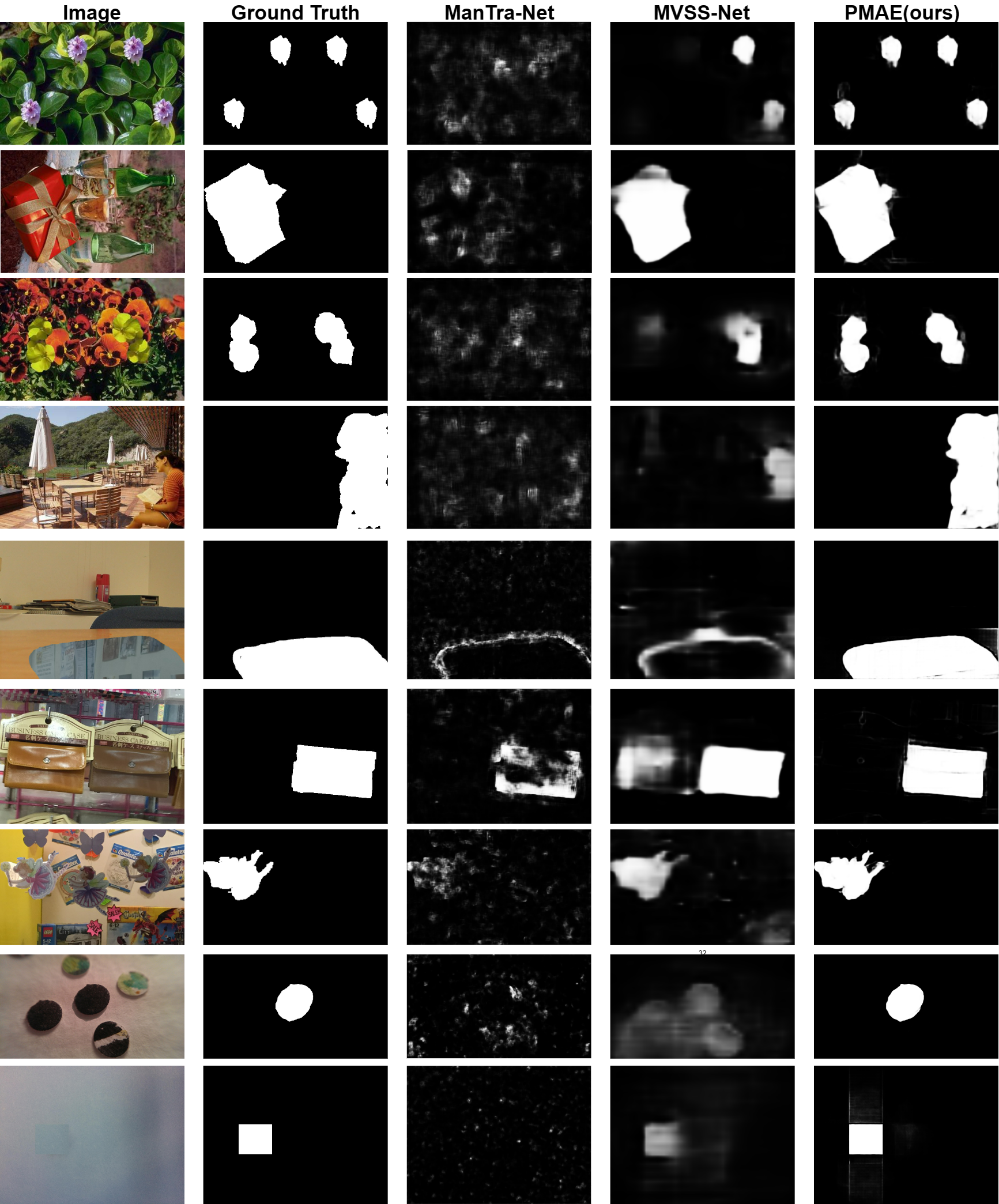}
  \caption[short]{Localization results of PMAE compared to various methods. All methods are trained on CASIAv2. Clear boundaries can observed in PMAE predictions.}
  \label{fig:visualize}
\end{figure}

\subsection{Compare with SOTA}
While there has been considerable research in the area of image manipulation localization, many works are not fully open-sourced. Some works that claim to be open-sourced do not provide access to their training code or huge private datasets, making it difficult to reproduce their results and compare them fairly with other models. In this section, we compare our model with state-of-the-art works that have published their models and parameters publicly. We evaluated the performance of these models on top-tier conferences and journals in recent years, following the commonly used protocol of training on the CASIAv2\cite{CASIA_2013} dataset and evaluating on smaller datasets.

\paragraph{Quantitative Analysis} We report the $F1$ score of these models and the complete results can be found in Table \ref{tab:sota}.  Some of the metrics reported in this section are referenced from MVSS-Net\cite{MVSS_2021}. We observe substantial performance improvements on our PMAE compared to previous works, even compared to the best-performing MVSS-Net++, PMAE still achieves an 18.2\% higher average $F1$ score of 0.502. This confirms the effectiveness of the proposed argument about the combination of multi-level vision traces. Although the F1 score of PMAE on COVERAGE dataset is slightly lower than that of MVSS-Net++, as we mentioned in section \ref{sec:ablation}, COVERAGE is a limited dataset with only one type of tampering and a small number of samples, so better performance on other larger datasets is more valuable and worth well. This suggests that PMAE has better generalization ability and indirectly demonstrates its effectiveness in discovering valuable tampering information.

\paragraph{Qualitative Analysis}
In Figure \ref{fig:visualize}, we present the predicted manipulation mask of our PMAE and compare it with the publicly available methods ManTra-Net\cite{Mantra_2019} and MVSS-Net\cite{MVSS_2021}. ManTra-Net and MVSS-Net are both FCN-based\cite{FCN_2015} IML methods utilizing handcrafted filters to extract low-level visual artifacts from images. However, we observed that although they can sometimes correctly detect areas with heavy artifacts (mainly boundaries of tampered areas), they are not confident in making a clear decision.
In contrast, PMAE, with the support of object-level semantics, can make a sharp and accurate prediction of tampered areas with clear boundaries. It can also effectively combine suspected dispersed regions into complete continuous areas.
\subsection{Robustness Evaluation}
\paragraph{Resize} As we applied a high-resolution input, we first evaluated the robustness of the PMAE toward Resize. However, most SoTA methods have not released their training code and the current convention does not emphasize the requirement for consistent resolution in comparing IML methods. Thus,  We resized images to the same resolution as the SoTA model and compared the performance of PMAE with them, results are shown in Table \ref{tab:rob_resize}.

\begin{table}[h]
\centering
  \caption{Robustness test toward resize algorithm.}
  \label{tab:rob_resize}
  \resizebox{1\columnwidth}{!}{
\begin{tabular}{@{}ccccccc@{}}
\toprule[2pt]
\textbf{Method}     & \textbf{Resize}      & \textbf{CASIAv1} & \textbf{Columbia} &\textbf{ Coverage} & \textbf{NIST16} & \textbf{MEAN}  \\ \midrule
GSR-Net    & 300×300     & 0.387   & 0.613    & 0.285    & 0.283  & 0.392 \\
PMAE       & 300×300     & 0.687   & 0.469    & 0.399    & 0.271  & 0.457 \\ \midrule
MVSS-Net++ & 512×512     & 0.513   & 0.660     & 0.482    & 0.304  & 0.490  \\
PMAE       & 512×512     & 0.649   & 0.736    & 0.441    & 0.300    & 0.531 \\ \midrule
PMAE       & Zero-padded & 0.688   & 0.860     & 0.473    & 0.311  & 0.583 \\ \bottomrule[2pt]
\end{tabular}
}
\end{table}
While it is natural for our model to experience some degree of performance decline, overall, it maintains a strong average performance level. The varying extent of decline across different datasets can be attributed to the fact that previous models directly extract specific features for identifying tampering, which are effective for specific tampering types. In contrast, PMAE primarily learns tampering features from the CASIAv2 dataset using self-supervised methods, making it more adaptable to homogeneous datasets.

\paragraph{Common Distortions} Additionally, following MVSS-Net\cite{mvsspp_2022}, we apply image distortion methods on raw input images from the CASIAv1 dataset and further evaluate the robustness of our PMAE model, utilizing pixel-level $F1$ score as the metrics to compare our model with ManTra-Net\cite{Mantra_2019} and MVSS-Net\cite{MVSS_2021}. Note that all methods are trained on pure CASIAv2 dataset without distortion. The distortion types include: 1) Gaussian blurring with a kernel size $k$; 2) JPEG compression with a quality factor $q$. The results are shown in Table \ref{tab:robustness}. 
The PMAE maintains relatively high performance against various compression methods, demonstrating the model's considerable robustness for practical applications.

\begin{table}[h]
  \begin{center}
    \caption{Robustness analysis of models on CASIAv1, evaluated with pixel-level $F1$ score(\%). }
  \label{tab:robustness}
  \begin{tabular}{l|c|c|c}
  \hline
  {Operations} & {ManTra-Net} & {MVSS-Net} & {PMAE(Ours)}  \\
  \hline\hline
  None  & 15.5 & 51.3 & \textbf{73.0}  \\
  \hline\hline
  JPEG Compress(100)  & 15.3 & 45.1 &  \textbf{{\color{red} 76.1$\uparrow $} }\\  
  JPEG Compress(90)   & 12.0 & 43.0 & \textbf{{\color{red}73.8$\uparrow $}}\\
  JPEG Compress(80)   & 8.1 & 42.4 & \textbf{68.2}\\
  JPEG Compress(70)   & 8.0 & 41.2 & \textbf{65.2}\\
  JPEG Compress(60)   & 8.7 & 40.1 & \textbf{62.0}\\
  JPEG Compress(50)   & 8.5 & 39.4 & \textbf{56.6}\\
  \hline\hline
  Gaussian Blur(size=5)   & 12.1 & 39.2 & \textbf{72.8}\\
  Gaussian Blur(size=11)   & 11.4 & 32.4 & \textbf{65.4}\\
  \hline
  \end{tabular}
  \end{center}

  \end{table}

However, an interesting phenomenon is that our model performs better at a compression quality of 100 and 90 for JPEG compression, which is unexpected. Therefore, we further test the robustness of our model on other datasets to explain this issue, results are shown in Table \ref{tab:ablation exception}.

\begin{table}[ht]
  \centering
    \caption{Exploration of abnormal increase after slight distortion on PAME. Evaluate with pixel-level $F1$ score, All anomalous growth is marked with the * symbol.}
  \label{tab:ablation exception}
  \resizebox{1\columnwidth}{!}{
  \begin{tabular}{@{}l|cccc@{}}
  \toprule[2pt]
  \textbf{Compression}           & \textbf{CASIAv1} & \textbf{NIST16} & \textbf{COVERAGE} & \textbf{Columbia} \\ \midrule
  None                  & 0.7307  & 0.3109 & 0.4731   & 0.8595   \\ \midrule
  JpegCompression(100)  & 0.7671*  & 0.2895 & 0.3907   & 0.8406   \\
  JpegCompression(90)   & 0.7435*  & 0.2742 & 0.3819   & 0.8044   \\ \midrule
  Gaussian Blur(size=3) & 0.7419*  & 0.3142* & 0.4054   & 0.8109   \\ 
  Gaussian Blur(size=5) & 0.7325*  & 0.3235* & 0.3789   & 0.7581   \\ 

  \bottomrule[2pt]
  \end{tabular}
  }

\end{table}

Here, we only observe this exception on NIST16 and CASIAv1 datasets. The explanation is as follows: there is a main commonality between these two datasets in our experiments, which is that they have undergone significant pre-processing. The CASIAv1 dataset itself resized all images to 256x384 (or 384x256) and added several noises before releasing, while NIST16, due to its large resolution, exceeded our 1024x1024 limit, so we downsampled them by resize. These operations have already destroyed a large number of low-level artifacts, such as noise from the camera, in these two datasets. In the model inference process, these datasets actually mostly rely on high-level visual features rather than low-level features to support their decisions. Therefore, slight blurring can help the model eliminate interference and better focus on object-level inconsistencies and incoherence, thus improving accuracy. In contrast, images in the COVERAGE and Columbia datasets are ``pure", without any pre-processing, which still contain a considerable amount of low-level information to support decision-making when passed to the model. Distortion will directly destroy this part of the information and reduce the total information available for supporting prediction, leading to decreased accuracy. Overall, this exceptional increment is also indirect evidence that our model has successfully reconciled multi-level visual information and can flexibly infer the manipulated area based on the richness of the two types of information.

\section{Conclusion}


This paper presents a novel approach to image manipulation detection by reformulating the task as a high-level vision task that greatly benefits from low-level features. Our proposed method, Perceptual Masked Autoencoder (PMAE), captures and balances multi-level visual information to effectively segment tampered areas. Through extensive experiments on multiple public datasets, PMAE has achieved state-of-the-art performance in F1-score, AUC, robustness, and generalization. Our results provide comprehensive evidence that incorporating both low-level and high-level features is necessary for effectively addressing image tampering, especially at the object level. 
\par 
In a nutshell, the proposed PMAE training paradigm represents a new state-of-the-art approach to solving multimedia image tampering. Future research should consider the distribution of low-level and high-level information in datasets and the real world when designing models. Furthermore, our proposed method can effectively address inpainting tampering, indicating its ability to recognize tampered information generated by large-scale models and its potential for practical applications.

\bibliographystyle{IEEEtran}
\bibliography{bibliography}

\vfill

\end{document}